\documentclass[10pt,twocolumn,letterpaper]{article}

\usepackage{cvpr}              
\usepackage{graphicx}
\usepackage{amsmath}
\usepackage{amssymb}
\usepackage{booktabs}
\usepackage{xcolor}
\usepackage{xspace}
\usepackage{multirow}
\usepackage{tablefootnote}
\usepackage{colortbl}
\usepackage{chngcntr}
\usepackage{inconsolata}
\newcounter{magicrownumbers}
\newcommand\rownumber{\stepcounter{magicrownumbers}\arabic{magicrownumbers}}

\usepackage{pifont}\newcommand{\cmark}{\ding{51}}\newcommand{\xmark}{\ding{55}}\definecolor{LightCyan}{rgb}{0.88,1,1}
\definecolor{LightGrey}{rgb}{0.88,0.88,0.88}

\newcommand{\demph}[1]{\textcolor{gray}{#1}}
\newcommand{\CMD}{PMD\xspace}
\newcommand{\FLAVA}{FLAVA\xspace}
\newcommand{\myparagraph}[1]{\vspace{0.25em}\noindent\textbf{#1}}

\usepackage[pagebackref,breaklinks,colorlinks]{hyperref}

\usepackage[capitalize]{cleveref}
\crefname{section}{Sec.}{Secs.}
\Crefname{section}{Section}{Sections}
\Crefname{table}{Table}{Tables}
\crefname{table}{Tab.}{Tabs.}

\def\cvprPaperID{****} \def\confName{CVPR}
\def\confYear{2022}

\begin{document}

\title{FLAVA: A Foundational Language And Vision Alignment Model}

\newcommand*\samethanks[1][\value{footnote}]{\footnotemark[#1]}
\author{
Amanpreet Singh\thanks{Equal contribution.} $\quad$ Ronghang Hu\samethanks $\quad$ Vedanuj Goswami\samethanks\\
\vspace{0.25em}
Guillaume Couairon $\quad$ Wojciech Galuba $\quad$ Marcus Rohrbach $\quad$ Douwe Kiela \\
\vspace{-0.5em}
Facebook AI Research (FAIR) \\
}
\maketitle

\vspace{-1em}
\begin{abstract}
\vspace{-0.5em}
State-of-the-art vision and vision-and-language models rely on large-scale visio-linguistic pretraining for obtaining good performance on a variety of downstream tasks. Generally, such models are often either cross-modal (contrastive) or multi-modal (with earlier fusion) but not both; and they often only target specific modalities or tasks. A promising direction would be to use a single holistic universal model, as a ``foundation’’, that targets all modalities at once---a true vision and language foundation model should be good at vision tasks, language tasks, and cross- and multi-modal vision and language tasks. We introduce FLAVA as such a model and demonstrate impressive performance on a wide range of 35 tasks spanning these target modalities.
\end{abstract}

\vspace{-1em}
\section{Introduction}

Large-scale pre-training of vision and language transformers has led to impressive performance gains in a wide variety of downstream tasks. In particular, contrastive methods such as CLIP \cite{radford2021learning} and ALIGN \cite{jia2021scaling} have shown that natural language supervision can lead to very high quality visual models for transfer learning.

Purely contrastive methods, however, also have important shortcomings. Their cross-modal nature does not make them easily usable on multimodal problems that require dealing with both modalities at the same time. They require large corpora, which for both CLIP and ALIGN have not been made accessible to the research community and the details of which remain shrouded in mystery, notwithstanding well-known issues with the construction of such datasets \cite{birhane2021multimodal}.

In contrast, the recent literature is rich with transformer models that explicitly target the multimodal vision-and-language domain by having earlier fusion and shared self-attention across modalities. For those cases, however, the unimodal vision-only or language-only performance of the model is often either glossed over or ignored completely.

If the future of our field lies in generalized ``foundation models'' \cite{bommasani2021opportunities} or ``universal'' transformers \cite{lu2021pretrained} with many different capabilities, then the following limitation should be overcome: a true foundation model in the vision and language space should not only be good at vision, or language, or vision-and-language problems--it should be good at all three, at the same time.

\begin{figure}[t]
\vspace{-1em}
\centering
\includegraphics[width=\linewidth]{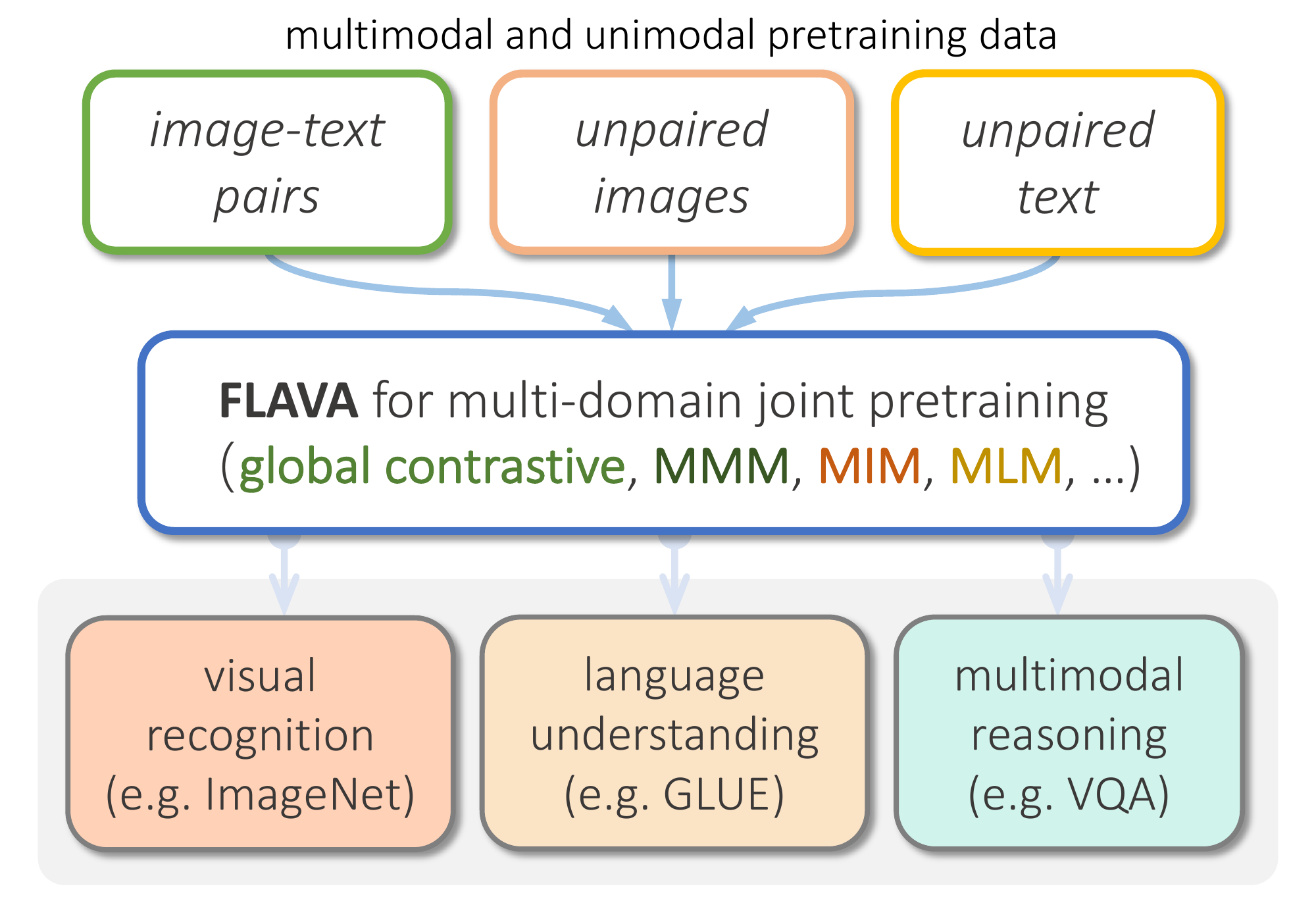}
\vspace{-2em}
\caption{We present FLAVA, a language and vision alignment model that learns strong representations from multimodal (image-text pairs) and unimodal data (unpaired images and text) and can be applied to target a broad scope of tasks from three domains (visual recognition, language understanding, and multimodal reasoning) under a common transformer model architecture.}
\label{fig:onecol}
\vspace{-1.5em}
\end{figure}

Combining information from different modalities into one universal architecture holds promise not only because it is similar to how humans make sense of the world, but also because it may lead to better sample efficiency and much richer representations.

\begin{table*}[t]
\vspace{-1.5em}
\small
\centering
\begin{tabular}{l|c@{}c@{\ \ }r|cccc|cccc}
\toprule
\multirow{2}[3]{*}{Method} &  \multicolumn{3}{c|}{\underline{Multimodal Pretraining data}} & \multicolumn{4}{c}{\underline{Pretraining Objectives}}  & \multicolumn{4}{|c}{\underline{Target Modalities}}  \\
 & public & dataset(s) & size & Contr. & ITM & Masking & Unimodal & V & CV\&L & MV\&L & L\\\midrule
CLIP \cite{radford2021learning} &  \xmark& WebImageText& 400M & \cmark & -- & -- & -- & \cmark & \cmark & -- & -- \\
ALIGN \cite{jia2021scaling} &  \xmark& JFT& 1.8B & \cmark & -- & -- & -- & \cmark & \cmark & -- & -- \\
SimVLM \cite{wang2021simvlm} &  \xmark & JFT &1.8B & -- & -- & PrefixLM & CLM & * & \cmark & \cmark & \cmark\\
UniT \cite{hu2021unit}  & -- & None & -- & -- & -- & -- & -- & * & -- & \cmark & \cmark\\
VinVL \cite{zhang2021vinvl} & \cmark &Combination & 9M & \cmark & -- & MLM & -- & -- & \cmark & \cmark & -- \\
ViLT \cite{kim2021vilt}  & \cmark & Combination & 10M & -- & \cmark & MLM & -- & -- & \cmark & \cmark & -- \\
ALBEF \cite{li2021albef}  & \cmark & Combination & 5M & \cmark & \cmark & MLM & -- & -- & \cmark & \cmark & -- \\
FLAVA (ours)  & \cmark&\CMD (Tbl.~\ref{tab:datasets}) &70M & \cmark & \cmark & MMM & MLM+MIM & \cmark & \cmark & \cmark & \cmark \\
\bottomrule
\end{tabular}
\vspace{-0.5em}
\caption{Comparison of recent models in different modalities. CV\&L and MV\&L stands for cross-modal and multi-modal vision-and-language. * means the modality is partially targeted (SimVLM \cite{wang2021simvlm} and UniT \cite{hu2021unit} include ImageNet and object detection, respectively).}
\label{tab:model_comparison}
\vspace{-1.5em}
\end{table*}

In this work, we introduce FLAVA, a foundational language and vision alignment model that explicitly targets vision, language, and their multimodal combination all at once. FLAVA learns strong representations through joint pretraining on both unimodal and multimodal data while encompassing cross-modal ``alignment'' objectives and multi-modal ``fusion'' objectives. We validate FLAVA by applying it to 35 tasks across vision, NLP, and multimodal domains and show impressive performance. An important advantage of our approach is that it was trained on a corpus of openly available datasets that is an order of magnitude smaller than datasets used in comparable models. Our models and code are available in \url{https://flava-model.github.io/}.

\section{Background}

The self-supervised pretraining paradigm has significantly advanced the state of the art across various domains, from natural language processing \cite{radford2021learning,radford2018improving,devlin2018bert,liu2019roberta,lewis2019bart,dong2019unilm,raffel2019t5,bao2020unilm2,conneau2019xlm,conneau2020xlmr,chi2019cross,chi2021infoxlm,chi2021xlme,ma2021deltalm}, to computer vision \cite{dosovitskiy2020image,touvron2020training,bao2021beit, carion2020end, gabeur2020multi, le2019multimodal, yuan2021temporal,bertasius2021space, neimark2021video, girdhar2019video, arnab2021vivit}, to speech recognition \cite{baevski2020wav2vec, conneau2020unsupervised, zhang2020pushing, liu2021tera, hsu2021hubert} and multimodal domains such as vision and language understanding \cite{jia2021scaling,carion2020end,hu2021unit,hu2020iterative,tan2019LXMERTLC,su2019vl,li2021albef, zhou2020unified, chen2020uniter, li2020oscar, li2021unimo, singh2020we, lu2019vilbert, li2019visualbert, lu202012, gan2020villa, zhang2021vinvl, wang2021simvlm, huang2021soho, yu2020ernie}. Even though this progress is based on a shared recipe of self-supervised learning on top of transformers, we are still missing major progress in building foundational models \cite{bommasani2021opportunities} that work well across all of these different domains and modalities at once.

Table~\ref{tab:model_comparison} shows an extensive comparison of popular and recent models \wrt our \FLAVA on multiple axes. Recent work either (i) focuses on a single target domain \cite{zhang2021vinvl, kim2021vilt}; (ii) targets a specific unimodal domain along with the joint vision-and-language domain \cite{jia2021scaling, radford2021learning}; or (iii) targets all domains but only a specific set of tasks in a particular domain.

SimVLM \cite{wang2021simvlm}, ALIGN \cite{jia2021scaling}, and CLIP \cite{radford2021learning} have demonstrated impressive gains by training transformer-based models on giant private paired image-and-text corpora, as opposed to the previous vision-and-language state-of-the-art such as VinVL \cite{zhang2021vinvl} and ViLT \cite{kim2021vilt}, which were trained on smaller public paired datasets \cite{conceptual_captions, ordonez2011sbu, coco, chen2015microsoft,visual_genome}.

Generally, models in the vision-and-language space can be divided into two categories: (i) dual encoders where the image and text are encoded separately followed by a shallow interaction layer for downstream tasks \cite{radford2021learning,jia2021scaling}; and (ii) fusion encoder(s) with self-attention spanning across the modalities \cite{hu2020iterative,tan2019LXMERTLC,su2019vl,li2021albef, zhou2020unified, chen2020uniter, li2020oscar, li2021unimo, lu2019vilbert, li2019visualbert, lu202012, gan2020villa, zhang2021vinvl, wang2021simvlm, huang2021soho}. The dual encoder approach works well for unimodal \cite{wang2019glue, wang2019superglue} and cross-modal retrieval tasks \cite{flickr30k,coco} but their lack of fusion usually causes them to underperform on tasks that involve visual reasoning and question answering \cite{goyal2017making,kiela2020hateful,singh2019towards,sidorov2020textcaps} which is where models based on fusion encoder(s) shine.

Within the fusion encoder category, a further distinction can be made as to whether the model uses a single transformer for early and unconstrained fusion between modalities (\eg, VisualBERT, UNITER, VLBERT, OSCAR \cite{li2019visualbert, chen2020uniter, su2019vl, li2020oscar, zhou2020unified}) or allows cross-attention only in specific co-attention transformer layers while having some modality specific layers (\eg, LXMERT, ViLBERT, ERNIE-ViL \cite{tan2019LXMERTLC, lu2019vilbert, lu202012, yu2020ernie}. Another distinguishing factor between different models lies in the image features that are used, ranging from region features \cite{lu2019vilbert,li2019visualbert,zhang2021vinvl}, to patch embeddings \cite{li2021albef,kim2021vilt,wang2021simvlm}, to convolution or grid features \cite{huang2020pixelbert,jiang2020defense}.

Dual encoder models use contrastive pretraining to predict the correct N paired combinations among N$^2$ possibilities. On the other hand, with fusion encoders, inspired by unimodal pretraining schemes such as masked language modeling \cite{devlin2018bert, liu2019roberta}, masked image modeling \cite{bao2021beit}, and causal language modeling \cite{radford2018improving}, numerous pretraining tasks have been explored: (i) Masked Language Modeling (MLM) for V\&L where masked words in the caption are predicted with help of the paired image \cite{li2019visualbert, lu2019vilbert, tan2019LXMERTLC}; (ii) prefixLM, where with the help of an image, the model tries to complete a caption \cite{wang2021simvlm,desai2020virtex}; (iii) image-text matching, where the model predicts whether given pair of image and text match or not; and (iv) masked region modeling, where the model regresses onto the image features or predicts its object class.

\begin{figure*}[t]
\vspace{-1.5em}
\centering
\includegraphics[width=0.95\linewidth]{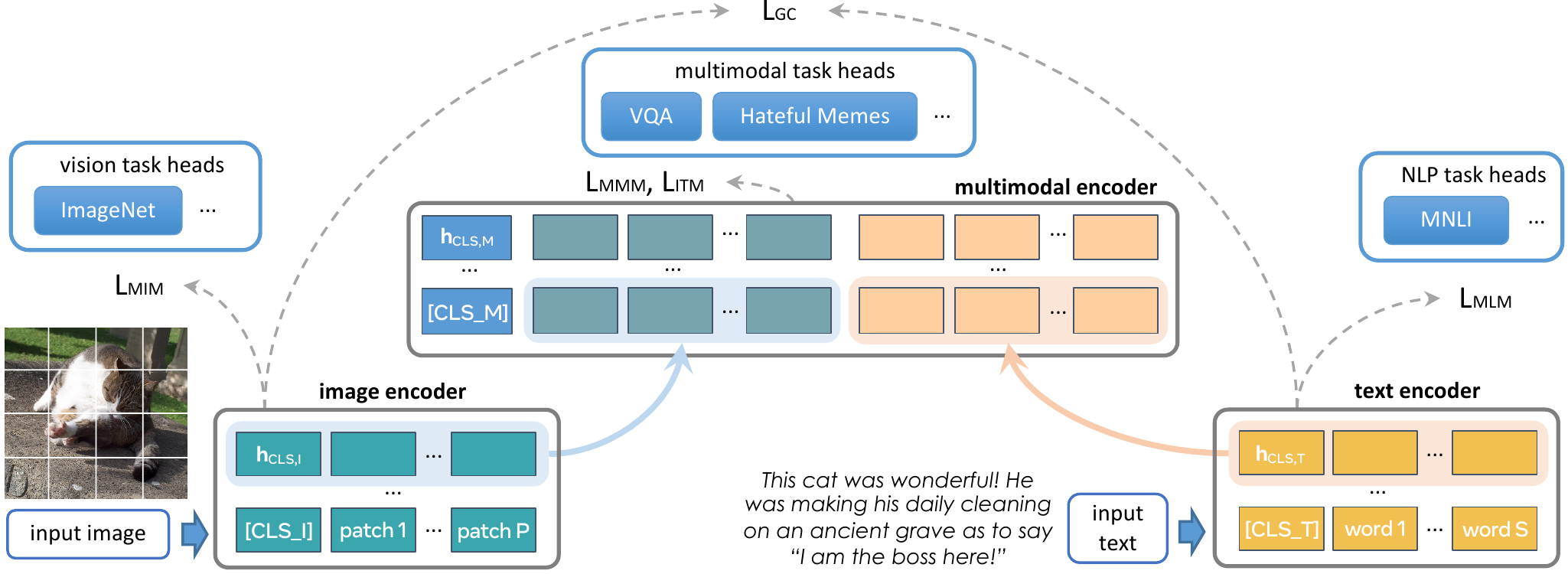}
\vspace{-.5em}
\caption{\textbf{An overview of our FLAVA model}, with an image encoder transformer to capture unimodal image representations, a text encoder transformer to process unimodal text information, and a multimodal encoder transformer that takes as input the encoded unimodal image and text and integrates their representations for multimodal reasoning. \textbf{During pretraining}, masked image modeling (MIM) and mask language modeling (MLM) losses are applied onto the image and text encoders over a single image or a text piece, respectively, while contrastive, masked multimodal modeling (MMM), and image-text matching (ITM) loss are used over paired image-text data. \textbf{For downstream tasks}, classification heads are applied on the outputs from the image, text, and multimodal encoders respectively for visual recognition, language understanding, and multimodal reasoning tasks.}
\label{fig:model}
\vspace{-1em}
\end{figure*}
Compared to previous work, our model \FLAVA works on a wide range of tasks in each of the vision, language, and vision-and-language domains. \FLAVA uses a shared trunk which was pretrained on only openly available public paired data. \FLAVA combines dual and fusion encoder approaches into one holistic model that can be pretrained with our novel \FLAVA pretraining scheme that leverages pretraining objectives from both categories. \FLAVA is designed to be able to take advantage of unpaired unimodal data along with multimodal paired data, resulting in a model that can handle unimodal and retrieval tasks as well as cross-modal and multi-modal vision-and-language tasks.

\section{FLAVA: A Foundational Language And Vision Alignment Model}

The goal of this work is to learn a foundational language and vision representation that enables unimodal vision and language understanding as well as multimodal reasoning, all within a single pre-trained model. We show how this can be achieved with a simple and elegant architecture based on transformers \cite{vaswani2017attention} (Sec.~\ref{sec:architecture}), which incorporates multimodal pretraining losses on image-text data (Sec.~\ref{sec:multimodal}) as well as unimodal pretraining losses on unimodal data (Sec.~\ref{sec:unimodal}). We discuss additional critical modeling insights in Sec.~\ref{sec:model:insights}. Finally, we demonstrate that our pretrained models can be successfully applied to a wide range of image, text, and multimodal tasks through both zero-shot and fine-tuning evaluations.

\subsection{The model architecture}
\label{sec:architecture}

The FLAVA model architecture is shown in Figure~\ref{fig:model}. The model involves an \textit{image encoder} to extract unimodal image representations, a \textit{text encoder} to obtain unimodal text representations, and a \textit{multimodal encoder} to fuse and align the image and text representations for multimodal reasoning, all of which are based on transformers.

\myparagraph{Image encoder.} We adopt the ViT architecture \cite{dosovitskiy2020image} for the image encoder. Given an input image, we resize it to a fixed image size and split the image into patches, which are then linearly embedded and fed into a transformer model (along with positional embeddings and an extra image classification token \texttt{[CLS\_I]}). The image encoder output is a list of image hidden state vectors $\{\mathbf{h}_I\}$, each corresponding to an image patch, plus an additional $\mathbf{h}_{\mathrm{CLS},I}$ for \texttt{[CLS\_I]}. We use the ViT-B/16 architecture for our image encoder.

\myparagraph{Text encoder.} Given an input piece of text (\eg, a sentence or a pair of sentences), we first tokenize and embed it into a list of word vectors following \cite{devlin2018bert}. Then, we apply a transformer model over the word vectors to encode them into a list of hidden state vectors $\{\mathbf{h}_T\}$, including $\mathbf{h}_{\mathrm{CLS},T}$ for the text classification \texttt{[CLS\_T]} token. Importantly, different from prior work, our text encoder has exactly the same architecture as the visual encoder, \ie, we use the same ViT architecture (but with different parameters) for both the visual and textual encoder, \ie ViT-B/16.

\myparagraph{Multimodal encoder.} We use a separate transformer to fuse the image and text hidden states. Specifically, we apply two learned linear projections over each hidden state vector in $\{\mathbf{h}_I\}$ and $\{\mathbf{h}_T\}$, and concatenate them into a single list with an additional \texttt{[CLS\_M]} token added, as shown in Figure~\ref{fig:model}. This concatenated list is fed into the multimodal encoder transformer (also based on the ViT architecture), allowing cross-attention between the projected unimodal image and text representations and fusing the two modalities. The output from the multimodal encoder is a list of hidden states $\{\mathbf{h}_M\}$, each corresponding to a unimodal vector from $\{\mathbf{h}_I\}$ or $\{\mathbf{h}_T\}$ (and a vector $\mathbf{h}_{\mathrm{CLS},M}$ for \texttt{[CLS\_M]}).

\myparagraph{Applying to downstream tasks.} The FLAVA model can be applied to both unimodal and multimodal tasks in a straightforward manner. For visual recognition tasks (\eg ImageNet classification), we apply a classifier head (\eg a linear layer or a multi-layer perceptron) on top of the unimodal $\mathbf{h}_{\mathrm{CLS},I}$ from the image encoder. Similarly, for language understanding and multimodal reasoning tasks, we apply a classifier head on top of $\mathbf{h}_{\mathrm{CLS},T}$ from the text encoder or $\mathbf{h}_{\mathrm{CLS},M}$ from the multimodal encoder, respectively. We pretrain the FLAVA model once, and evaluate it separately on each downstream task. More details about finetuning, linear, and zero-shot evaluation on specific tasks can be found in the supplemental.

\subsection{Multimodal pretraining objectives}
\label{sec:multimodal}

We aim to obtain strong representations through pretraining on both multimodal data (paired image and text) as well as unimodal data (unpaired images or text). FLAVA pretraining involves the following multimodal objectives.

\myparagraph{Global contrastive (GC) loss.} Our image-text contrastive loss resembles that of CLIP \cite{radford2021learning}. Given a batch of images and text, we maximize the cosine similarities between matched image and text pairs and minimize those for the unmatched pairs. This is accomplished by linearly projecting each $\mathbf{h}_{\mathrm{CLS},I}$ and $\mathbf{h}_{\mathrm{CLS},T}$ into an embedding space, followed by L2-normalization, dot-product, and a softmax loss scaled by temperature.

Large models are often trained using multiple GPUs data parallelism, where the samples in a batch are split across GPUs. When gathering embeddings for the image and text contrastive objective, the open-source CLIP implementation \cite{ilharco_gabriel_2021_5143773} only back-propagates the gradients of the contrastive loss to the embeddings from the local GPU where the dot-product is performed. In contrast, through experiments that can be found in the supplemental, we observe a noticeable performance gain by performing full back-propagation across GPUs compared to only doing back-propagation locally. We call our loss ``global contrastive'' $L_{\mathrm{GC}}$ to distinguish it from ``local contrastive'' approaches.

\myparagraph{Masked multimodal modeling (MMM).} While a number of previous vision-and-language pretraining approaches (\eg \cite{li2019visualbert}) have focused on masked modeling of the text modality by reconstructing masked tokens from the multimodal input, most of them do not involve masked learning on image modality directly at the image pixel level in an end-to-end manner. Here, we introduce a novel masked multimodal modeling (MMM) pretraining objective $L_{\mathrm{MMM}}$ that masks both the image patches and the text tokens and jointly works on both modalities.

Specifically, given an image and text input, we first tokenize the input image patches using a pretrained dVAE tokenizer \cite{dalle}, which maps each image patch into an index in a visual codebook similar to a word dictionary (we use the same dVAE tokenizer as in \cite{bao2021beit}). Then, we replace a subset of image patches based on rectangular block image regions following BEiT \cite{bao2021beit} and 15\% of text tokens following BERT \cite{devlin2018bert} with a special \texttt{[MASK]} token. Then, from the multimodal encoder's output $\{\mathbf{h}_M\}$, we apply a multi-layer perceptron to predict the visual codebook index of the masked image patches, or the word vocabulary index of the masked text tokens.

This objective can be seen as an extension of the multimodal masked language modeling such that it incorporates masking on the image side. In our experiments, we find that our MMM pretraining leads to improvements over and in addition to the contrastive loss pretraining, especially for multimodal downstream tasks such as VQA. Note that we apply global contrastive loss on image patches and text tokens without any masking, which are forwarded through the image and text encoders separately from the MMM loss.

\myparagraph{Image-text matching (ITM).} Finally, we add an image-text matching loss $L_{\mathrm{ITM}}$ following prior vision-and-language pretraining literature\cite{lu2019vilbert,tan2019LXMERTLC,chen2020uniter}. During pretraining, we feed a batch of samples including both matched and unmatched image-text pairs. Then, on top of $\mathbf{h}_{\mathrm{CLS},M}$ from the multimodal encoder, we apply a classifier to decide if an input image and text match each other.

\begin{figure*}[t]
\vspace{-1.5em}
\centering
\includegraphics[width=0.95\linewidth]{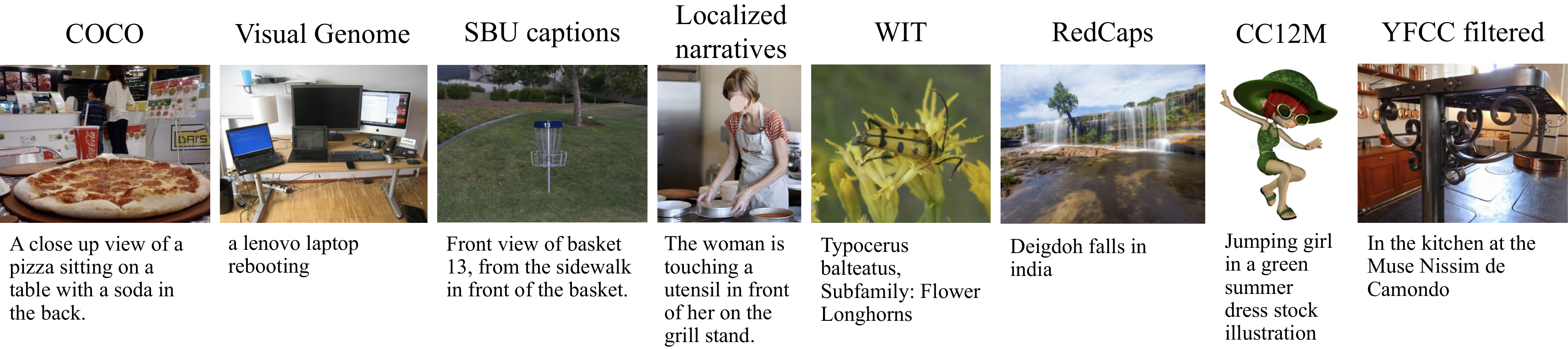}
\vspace{-0.9em}
\caption{Representative examples from various subsets of our pretraining dataset (details in Sec.~\ref{sec:dataset}).}
\label{fig:cmd}
\vspace{-1em}
\end{figure*}

\subsection{Unimodal pretraining objectives}
\label{sec:unimodal}

While the objectives in Sec.~\ref{sec:multimodal} allow pretraining the FLAVA model on paired image-and-text data, the vast majority of datasets (such as ImageNet for images and CC-News for text) are unimodal without paired data from the other modality. To efficiently learn a representation for a wide range of downstream tasks, we would also like to leverage these datasets and incorporate unimodal and unaligned information into our representations.

In this work, we introduce knowledge and information from these unimodal datasets through 1) pretraining the image encoder and text encoder on unimodal datasets; 2) pretraining the entire FLAVA model jointly on \textit{both} unimodal and multimodal datasets; or 3) a combination of both by starting from pretrained encoders and then jointly training. When applied to stand-alone image or text data, we adopt masked image modeling (MIM) and masked language modeling (MLM) losses over the image and text encoders respectively, as described in what follows.

\myparagraph{Masked image modeling (MIM).} On unimodal image datasets, we mask a set of image patches following the rectangular block-wise masking in BEiT \cite{bao2021beit} and reconstruct them from other image patches. The input image is first tokenized using a pretrained dVAE tokenizer \cite{dalle} (same as the one used in the MMM objective in Sec.~\ref{sec:multimodal}), and then a classifier is applied on the image encoder outputs $\{\mathbf{h}_I\}$ to predict the dVAE tokens of the masked patches.

\myparagraph{Masked language modeling (MLM).} We apply a masked language modeling loss \cite{devlin2018bert} on top of the text encoder to pretrain on stand-alone text datasets. A fraction (15\%) of the text tokens are masked in the input, and reconstructed from the other tokens using a classifier over the unimodal text hidden states output $\{\mathbf{h}_T\}$.

\myparagraph{Encoder initialization from unimodal pretraining.} We use three sources of data for pretraining: unimodal image data (ImageNet-1K\cite{imagenet}), unimodal text data (CCNews\cite{liu2019roberta} and BookCorpus\cite{zhu2015aligning}), and multimodal image-text paired data (Sec.~\ref{sec:dataset}). We first pretrain the text encoder with the MLM objective on the unimodal text dataset. We experiment with different ways for pretraining the image encoder: we pretrain the image encoder on unpaired image datasets with either MIM or the DINO objective \cite{caron2021emerging}, before joint training on both unimodal and multimodal datasets. We empirically found the latter to work quite well, despite the switch to an MIM objective on images post-initialization (more details in supplemental).
Then, we initialize the whole FLAVA model with the two respective unimodally-pretrained encoders, or when we train from scratch, we initialize randomly. We always initialize the multimodal encoder randomly for pretraining.

\myparagraph{Joint unimodal and multimodal training.} After unimodal pretraining of the image and text encoders, we continue training the entire FLAVA model jointly on the three types of datasets with round-robin sampling. In each training iteration, we choose one of the datasets according to a sampling ratio that we determine empirically (see supplemental) and obtain a batch of samples. Then, depending on the dataset type, we apply unimodal MIM on image data, unimodal MLM on text data, or the multimodal losses (contrastive, MMM, and ITM) in Sec.~\ref{sec:multimodal} on image-text pairs.

\subsection{Implementation details}
\label{sec:model:insights}

We find that the optimizer hyperparameters play a critical role in effective pretraining. A large batch size, a large weight decay, and a long warm-up are all important for preventing divergence with a large learning rate (we use 8,192 batch size, 1e-3 learning rate, 0.1 weight decay, and 10,000 iteration warm-up in our pretraining tasks together with the AdamW optimizer \cite{loshchilov2019decoupled,kingma2014adam}). In addition, the ViT transformer architecture (which applies layer norm \cite{ba2016layernorm} \emph{before} the multi-head attention rather than after \cite{xiong2020layer}) provides more robust learning for the text encoder under large learning rate than the BERT \cite{devlin2018bert} transformer architecture. FLAVA is implemented using the open-source MMF \cite{singh2020mmf} and fairseq \cite{ott2019fairseq} libraries. We use Fully-Sharded Data Parallel (FSDP) \cite{rajbhandari2019zero,rajbhandari2021zero} and train in full FP16 precision except the layer norm \cite{ba2016layernorm} to reduce GPU memory consumption.

\subsection{Data: Public Multimodal Datasets (\CMD)}
\label{sec:dataset}

\begin{table}[t]
\small
\resizebox{\columnwidth}{!}{
\begin{tabular}{lcc}
\toprule
& \shortstack{\#Image-Text Pairs} & \shortstack{Avg. text length} \\ 
\midrule
COCO \cite{coco} & 0.9M & 12.4 \\
SBU Captions \cite{ordonez2011sbu} & 1.0M  & 12.1 \\
Localized Narratives \cite{localized_narratives} &  1.9M & 13.8 \\
Conceptual Captions \cite{conceptual_captions} & 3.1M & 10.3\\
Visual Genome \cite{visual_genome} & 5.4M & 5.1\\
Wikipedia Image Text \cite{wikipedia_image_text}  & 4.8M & 12.8\\
Conceptual Captions 12M \cite{conceptual_captions_12M}  & 11.0M& 17.3\\
Red Caps \cite{redcaps} &  11.6M& 9.5\\
YFCC100M  \cite{yfcc100m}, filtered & 30.3M & 12.7\\
\midrule
Total & 70M & 12.1 \\ 
\bottomrule
\end{tabular}}
\vspace{-0.5em}
\caption{Public Multimodal Datasets (\CMD) corpus used in FLAVA multimodal pretraining, which consists of publicly available datasets with a total size of 70M image and text pairs.}
\label{tab:datasets}
\vspace{-1em}
\end{table}

For multimodal pretraining, we constructed a corpus out of publicly available sources of image-text data, which are presented in Table \ref{tab:datasets} with examples in Fig.~\ref{fig:cmd}. The total count of text-image pairs is 70M, including 68M unique images, and the average caption length is 12.1 words. For the YFCC100M dataset \cite{yfcc100m}, we filter the image-text data by discarding non-English captions and only keeping captions that contain more than two words. We first consider the \textit{description} field of each image, if this does not pass our filters we consider the \textit{title} field. Other than that, we did not do any additional filtering. Importantly, this corpus entirely consists of open datasets that are freely accessible by other researchers, facilitating reproducibility and enabling future work by the community.

\section{Experiments}

We evaluate FLAVA across vision, language, and multimodal tasks. For vision, we evaluate on 22 common vision tasks. For NLP, we evaluate on 8 tasks from the GLUE \cite{wang2019glue} benchmark. For multimodal, we evaluate on VQAv2 \cite{goyal2017making}, SNLI-VE \cite{xie2019visual}, Hateful Memes \cite{kiela2020hateful}, as well as Flickr30K \cite{flickr30k} and COCO \cite{coco} image and text retrieval.

We compare our joint pretraining method ($\mathrm{FLAVA}$ in Table \ref{tab:main_table} and \ref{tab:ablations}) with other settings, on this diverse array of 35 tasks. We report the average performance on the NLP, vision, and multimodal tasks, and an additional macro average across all the three modalities in Table \ref{tab:main_table}, and also the detailed the performance on each task in Table \ref{tab:ablations}.

\begin{table}[t]
\setcounter{magicrownumbers}{0}
\resizebox{\columnwidth}{!}{
\begin{tabular}{lccc|c}
\toprule
 & Vision & NLP & Multi-modal & Macro \\
Method & Avg. & Avg. & Avg. & Avg. \\
\midrule
\texttt{\rownumber} MIM & 57.46 & -- & -- & 19.15 \\
\texttt{\rownumber} MLM & -- & 71.55 & -- & 23.85   \\
\texttt{\rownumber} $\mathrm{FLAVA}_\mathrm{C}$ & 64.80 & 79.14  & 66.25 & 70.06 \\
\texttt{\rownumber} $\mathrm{FLAVA}_\mathrm{MM}$ & 74.22 & 79.35  & 69.11 & 74.23 \\
\texttt{\rownumber} $\mathrm{FLAVA}$ w/o unimodal init & 75.55 & 78.29 & 67.32 & 73.72 \\
\rowcolor{LightGrey}
\texttt{\rownumber} $\mathrm{FLAVA}$ & \textbf{78.19} &\textbf{ 79.44} & \textbf{69.92} & \textbf{75.85} \\\bottomrule
\end{tabular}}
\vspace{-.5em}
\caption{Our full $\mathrm{FLAVA}$ pretraining (row 6) achieves the best average scores on vision, language, and multimodal tasks compared to ablations. Row 1 to 4 are pretrained on PMD while row 5 and 6 also involve unimodal IN-1k, CCNews, and BookCorpus datasets.}
\label{tab:main_table}
\vspace{-1em}
\end{table}

\begin{table*}[t]
\vspace{-1.5em}
\centering
\footnotesize
\begin{tabular}{l@{\ }c|ccccccc|c}
\toprule
&  & \multicolumn{1}{c}{MIM} & \multicolumn{1}{c}{MLM} &
\multicolumn{1}{c}{$\mathrm{FLAVA}_\mathrm{C}$} & \multicolumn{1}{c}{$\mathrm{FLAVA}_\mathrm{MM}$}  & \multicolumn{1}{c}{$\mathrm{FLAVA}$ w/o init} & \multicolumn{1}{c}{$\mathrm{FLAVA}$} & \multicolumn{1}{c|}{CLIP} & \multicolumn{1}{c}{CLIP} \\
&  & \small\texttt{1} & \small\texttt{2} & \small\texttt{3} & \small\texttt{4} & \small\texttt{5} & \small\texttt{6} & \small\texttt{7} & \small\texttt{8}   \\
\midrule 
Datasets & Eval method &  PMD & PMD &  PMD & PMD & \multicolumn{2}{c}{(PMD+IN-1k+CCNews+BC)} & PMD & 400M\cite{radford2021learning} \\
\midrule 
MNLI \cite{williams2018broad} & fine-tuning & -- & 73.23 & 70.99 & 76.82 & 78.06 & \underline{\textbf{80.33}} & 32.85 & 33.52 \\
CoLA \cite{warstadt2019neural} & fine-tuning & -- & 39.55 & 17.58 & 38.97 & 44.22 & \underline{\textbf{50.65}} & 11.02 & 25.37 \\
MRPC \cite{mrpc2005} & fine-tuning & -- & 73.24 & 76.31 & 79.14 & 78.91 & \underline{\textbf{84.16}} & 68.74 & 69.91 \\
QQP \cite{qqp} & fine-tuning & -- & 86.68 & 85.94 & 88.49 & 98.61 & \underline{\textbf{88.74}} & 59.17 & 65.33 \\
SST-2 \cite{sst2013} & fine-tuning & -- & 87.96 & 86.47 & 89.33 & 90.14 & \underline{\textbf{90.94}} & 83.49 & 88.19 \\
QNLI \cite{rajpurkar2016squad} & fine-tuning & -- & 82.32 & 71.85 & 84.77 & 86.40 & \underline{\textbf{87.31}} & 49.46 & 50.54 \\
RTE \cite{dagan2006pascal,bar2006second,giampiccolo2007third,bentivogli2009fifth} & fine-tuning & -- & 50.54 & 51.99 & 51.99 & 54.87 & \underline{\textbf{57.76}} & 53.07 & 55.23 \\
STS-B \cite{agirre2007semantic} & fine-tuning & -- & 78.89 & 57.28 & 84.29 & 83.21 & \underline{\textbf{85.67}} & 13.70 & 15.98 \\
\midrule 
\textbf{NLP Avg.} &  & -- & 71.55 & 64.80 & 74.22 & 75.55 & \underline{\textbf{78.19}} & 46.44 & 50.50 \\
\midrule 
ImageNet \cite{imagenet} & linear eval & 41.79 & -- & 74.09 & 74.34 & 73.49 & \textbf{75.54} & 72.95 & \underline{80.20} \\
Food101 \cite{bossard14} & linear eval & 53.30 & -- & 87.77 & 87.53 & 87.39 & \textbf{88.51} & 85.49 & \underline{91.56} \\
CIFAR10 \cite{krizhevsky2009learning} & linear eval & 76.20 & -- & \textbf{93.44} & 92.37 & 92.63 & 92.87 & 91.25 & \underline{94.93} \\
CIFAR100 \cite{krizhevsky2009learning} & linear eval & 55.57 & -- & \textbf{78.37} & 78.01 & 76.49 & 77.68 & 74.40 & \underline{81.10} \\
Cars \cite{KrauseStarkDengFei-Fei_3DRR2013} & linear eval & 14.71 & -- & \textbf{72.12} & 72.07 & 66.81 & 70.87 & 62.84 & \underline{85.92} \\
Aircraft \cite{maji13fine-grained} & linear eval & 13.83 & -- & \textbf{49.74} & 48.90 & 44.73 & 47.31 & 40.02 & \underline{51.40} \\
DTD \cite{cimpoi14describing} & linear eval & 55.53 & -- & 76.86 & 76.91 & 75.80 & \textbf{77.29} & 73.40 & \underline{78.46} \\
Pets \cite{parkhi12a} & linear eval & 34.48 & -- & \textbf{84.98} & 84.93 & 82.77 & 84.82 & 79.61 & \underline{91.66} \\
Caltech101 \cite{FeiFei2004LearningGV} & linear eval & 67.36 & -- & 94.91 & 95.32 & 94.95 & \underline{\textbf{95.74}} & 93.76 & 95.51 \\
Flowers102 \cite{Nilsback08} & linear eval & 67.23 & -- & 96.36 & \textbf{96.39} & 95.58 & 96.37 & 94.94 & \underline{97.12} \\
MNIST \cite{lecun2010mnist} & linear eval & 96.40 & -- & 98.39 & 98.58 & \textbf{98.70} & 98.42 & 97.38 & \underline{99.01} \\
STL10 \cite{coates2011analysis} & linear eval & 80.12 & -- & 98.06 & 98.31 & 98.32 & \textbf{98.89} & 97.29 & \underline{99.09} \\
EuroSAT \cite{helber2019eurosat} & linear eval & 95.48 & -- & 97.00 & 96.98 & 97.04 & \underline{\textbf{97.26}} & 95.70 & 95.38 \\
GTSRB \cite{stallkamp2011german} & linear eval & 63.14 & -- & 78.92 & 77.93 & 77.71 & \textbf{79.46} & 76.34 & \underline{88.61} \\
KITTI \cite{geiger2013vision} & linear eval & 86.03 & -- & 87.83 & 88.84 & 88.70 & \underline{\textbf{89.04}} & 84.89 & 86.56 \\
PCAM \cite{veeling2018rotation} & linear eval & 85.10 & -- & 85.02 & 85.51 & \underline{\textbf{85.72}} & 85.31 & 83.99 & 83.72 \\
UCF101 \cite{soomro2012ucf101} & linear eval & 46.34 & -- & 82.69 & 82.90 & 81.42 & \textbf{83.32} & 77.85 & \underline{85.17} \\
CLEVR \cite{johnson2017clevr} & linear eval & 61.51 & -- & 79.35 & \underline{\textbf{81.66}} & 80.62 & 79.66 & 73.64 & 75.89 \\
FER 2013 \cite{goodfellow2014explaining} & linear eval & 50.98 & -- & 59.96 & 60.87 & 58.99 & \textbf{61.12} & 57.04 & \underline{68.36} \\
SUN397 \cite{xiao2016sun} & linear eval & 52.45 & -- & 81.27 & 81.41 & 81.05 & \underline{\textbf{82.17}} & 79.96 & 82.05 \\
SST \cite{radford2021learning} & linear eval & 57.77 & -- & 56.67 & \textbf{59.25} & 56.40 & 57.11 & 56.84 & \underline{74.68} \\
Country211 \cite{radford2021learning} & linear eval & 8.87 & -- & 27.27 & 26.75 & 27.01 & \textbf{28.92} & 25.12 & \underline{30.10} \\
\midrule 
\textbf{Vision Avg.} &  & 57.46 & -- & 79.14 & 79.35 & 78.29 & \textbf{79.44} & 76.12 & \underline{82.57} \\
\midrule
VQAv2 \cite{goyal2017making} & fine-tuning & -- & -- & 67.13 & 71.69 & 71.29 & \underline{\textbf{72.49}} & 59.81 & 54.83 \\
SNLI-VE \cite{xie2019visual} & fine-tuning & -- & -- & 73.27 & 78.36 & 78.14 & \underline{\textbf{78.89}} & 73.53 & 74.27 \\
Hateful Memes \cite{kiela2020hateful} & fine-tuning & -- & -- & 55.58 & 70.72 & \underline{\textbf{77.45}} & 76.09 & 56.59 & 63.93 \\
Flickr30K \cite{flickr30k} TR R@1 & zero-shot & -- & -- & 68.30 & \textbf{69.30} & 64.50 & 67.70 & 60.90 & \underline{82.20} \\
Flickr30K \cite{flickr30k} TR R@5 & zero-shot & -- & -- & 93.50 & 92.90 & 90.30 & \textbf{94.00} & 88.90 & \underline{96.60} \\
Flickr30K \cite{flickr30k} IR R@1 & zero-shot & -- & -- & 60.56 & 63.16 & 60.04 & \underline{\textbf{65.22}} & 56.48 & 62.08 \\
Flickr30K \cite{flickr30k} IR R@5 & zero-shot & -- & -- & 86.68 & 87.70 & 86.46 & \underline{\textbf{89.38}} & 83.60 & 85.68 \\
COCO \cite{coco} TR R@1 & zero-shot & -- & -- & 43.08 & \textbf{43.48} & 39.88 & 42.74 & 37.12 & \underline{52.48} \\
COCO \cite{coco} TR R@5 & zero-shot & -- & -- & 75.82 & \underline{\textbf{76.76}} & 72.84 & \underline{\textbf{76.76}} & 69.48 & 76.68 \\
COCO \cite{coco} IR R@1 & zero-shot & -- & -- & 37.59 & \underline{\textbf{38.46}} & 34.95 & 38.38 & 33.29 & 33.07 \\
COCO \cite{coco} IR R@5 & zero-shot & -- & -- & 67.28 & \underline{\textbf{67.68}} & 64.63 & 67.47 & 62.47 & 58.37 \\
\midrule
\textbf{Multimodal Avg.} &  & -- & -- & 66.25 & 69.11 & 67.32 & \underline{\textbf{69.92}} & 62.02 & 67.29 \\
\midrule
\textbf{Macro Avg.} &  & 19.15 & 23.85 & 70.06 & 74.23 & 73.72 & \underline{\textbf{75.85}} & 61.52 & 66.78 \\
\bottomrule
\end{tabular}
\vspace{-.5em}
\caption{\textbf{Comparing our full FLAVA pretraining with other settings}, where $\mathrm{FLAVA}$ gets the highest macro average score. MNLI numbers are average of MNLI-m and MNLI-mm. MRPC and QQP numbers are average of accuracy and F1. We report PCC for CoLA, MCC for STS-B, and AUROC for Hateful Memes, respectively. We perform zero-shot text retrieval and image retrieval (TR and IR) on Flickr30K and COCO based on their matching scores from the contrastive loss and report top-1 and top-5 recall. For all other tasks we report accuracy. Column 8 is the best released model in \cite{radford2021learning} based on ViT-B/16 pretrained on 400M image-text pairs. The overall best result is \underline{underlined} while \textbf{bold} signifies the best on public data (PMD and unimodal).}
\label{tab:ablations}
\vspace{-1em}
\end{table*}

\begin{table*}[t]
\vspace{-1.5em}
\setcounter{magicrownumbers}{0}
\footnotesize
\centering
\setlength{\tabcolsep}{5pt}
\begin{tabular}{ccl|c@{\ \ }c@{\ }c|c@{\ \ }c@{\ \ }cccccc|c}
\toprule
& \multicolumn{1}{@{}c@{}}{public}& & \multicolumn{3}{c|}{Multimodal Tasks} & \multicolumn{8}{c|}{Language Tasks} & ImageNet \\
& \multicolumn{1}{@{}c@{}}{data}& & VQAv2 & SNLI-VE & HM  & CoLA & SST-2 & RTE & MRPC & QQP & MNLI & QNLI & STS-B & linear eval \\
\midrule
\demph{\texttt{\rownumber}} & \demph{\checkmark}& \demph{BERT$_\text{base}$} \cite{devlin2018bert} & \demph{--} & \demph{--} & \demph{--} & \demph{54.6} & \demph{92.5} & \demph{62.5} & \demph{81.9/87.6} & \demph{90.6/87.4} & \demph{84.4} & \demph{91.0} & \demph{88.1}  & \demph{--} \\
\midrule
\texttt{\rownumber}  & \xmark & CLIP-ViT-B/16 \cite{radford2021learning} & 55.3 & 74.0~~ & 63.4~~ & 25.4 & 88.2 & 55.2 & 74.9/65.0 & 76.8/53.9 & 33.5 & 50.5 & 16.0 & 80.2 \\
\texttt{\rownumber} & \xmark&SimVLM$_\text{base}$ \cite{wang2021simvlm} & \underline{77.9} & \underline{84.2}~~ & -- & 46.7 & 90.9 & \underline{63.9} & 75.2/84.4 & \underline{90.4/87.2} & \underline{83.4} & \underline{88.6} & -- & \underline{80.6} \\
\midrule
\texttt{\rownumber} & \checkmark&VisualBERT \cite{li2019visualbert} & 70.8 & 77.3$^\dagger$ & 74.1$^\ddag$ & 38.6 & 89.4 & 56.6 & 71.9/82.1 & 89.4/86.0 & \bf  81.6 & 87.0 & 81.8 & -- \\
\texttt{\rownumber} & \checkmark&UNITER$_\text{base}$ \cite{chen2020uniter} & 72.7 & 78.3~~ & --~ & 37.4 & 89.7 & 55.6 & 69.3/80.3 & 89.2/85.7 & 80.9 & 86.0 & 75.3 & -- \\
\texttt{\rownumber} & \checkmark&VL-BERT$_\text{base}$ \cite{su2019vl} & 71.2 & -- & --~ & 38.7 & 89.8 & 55.7 & 70.6/81.8 & 89.0/85.4 & 81.2 & 86.3 &  82.9 & -- \\
\texttt{\rownumber} & \checkmark&ViLBERT \cite{lu2019vilbert} & 70.6 & 75.7$^\dagger$ & 74.1$^\ddag$ & 36.1 & 90.4 & 53.7 & 69.0/79.4 & 88.6/85.0 & 79.9 & 83.8  & 77.9 & -- \\
\texttt{\rownumber} & \checkmark&LXMERT \cite{tan2019LXMERTLC} & 72.4 & -- & --~ & 39.0 & 90.2 & 57.2 & 69.7/80.4 & 75.3/75.3 & 80.4 & 84.2 & 75.3 & -- \\
\texttt{\rownumber} & \checkmark&UniT \cite{hu2021unit} & 67.0 & 73.1~~ & --~ & -- & 89.3 & -- & -- & 90.6/ ~~--~~ & 81.5 & \bf  88.0 & -- & -- \\
\texttt{\rownumber} & \checkmark&CLIP-ViT-B/16 (PMD) & 59.8 & 73.5~~ & 56.6~~ & 11.0 & 83.5 & 53.1 & 63.5/68.7 & 75.4/43.0 & 32.9 & 49.5 & 13.7 & 73.0 \\
\rowcolor{LightGrey}
\texttt{\rownumber} & \checkmark&FLAVA (ours) & \bf 72.8 & \bf 79.0~~ & \underline{\bf 76.7}~~ & \underline{\bf 50.7} & \underline{\bf 90.9} & \bf  57.8 & \underline{\bf 81.4/86.9} & \underline{\bf 90.4/87.2} &  80.3 & 87.3 & \underline{\bf  85.7} & \bf  75.5 \\
\bottomrule
\end{tabular}
\vspace{-.5em}
\caption{\textbf{Comparing FLAVA (Table~\ref{tab:ablations} column 6) with previous models on multimodal tasks, language tasks, and ImageNet linear evaluation.} We report results on development sets of the GLUE benchmark \cite{wang2019glue}. We report Matthew's Correlation for CoLA; accuracy/F1 for MRPC and QQP; the Pearson/Spearman correlation for STS-B; average of mismatched and matched accuracy for MNLI; AUROC for Hateful Memes; test-dev VQA score for VQAv2 and accuracy for all other tasks. The results for BERT and other VLP methods on GLUE benchmark are obtained from \cite{iki2021effect}. The results on V\&L tasks are from original papers. For UniT, we use ``shared, (COCO init.)'' version. Note that SimVLM is pretrained on an order of magnitude more data than FLAVA (1.8B vs 70M). $^\dagger$: taken from \cite{singh2020we}; $^\ddag$: taken from \cite{kiela2020hateful}.
The overall best result among the multimodal approaches is \underline{underlined} while \textbf{bold} signifies the best model trained on public data.}
\label{tab:sota_comp}
\vspace{-1em}
\end{table*}

\myparagraph{Full FLAVA pretraining achieves the best results.} Table~\ref{tab:main_table}
shows baselines and different ablation settings of FLAVA, including: models trained with unimodal MIM and MLM losses, $\mathrm{FLAVA}_\mathrm{C}$ trained with only image-text contrastive loss, $\mathrm{FLAVA}_\mathrm{MM}$ trained only on multimodal data, models without unimodal initialization, and the full model (each setting is detailed in the paragraphs below). The full FLAVA model in row 6 outperforms all other settings in average performance over NLP, vision, and multimodal tasks.

\myparagraph{Effective global contrastive loss in FLAVA.} We next perform a step-by-step ablation of our model (Table \ref{tab:ablations}). We first train a restricted version of FLAVA using only the global contrastive loss $L_{\mathrm{GC}}$ in Sec.~\ref{sec:multimodal} on multimodal data, denoted as $\mathrm{FLAVA}_\mathrm{C}$ in column 3. This restricted setting is a conceptually similar model to CLIP \cite{radford2021learning} that also involves a contrastive loss, and we compare against the CLIP model trained on the same PMD data with the same ViT-B/16 image encoder as a baseline (using the open-source implementation in \cite{ilharco_gabriel_2021_5143773}), denoted as $\mathrm{CLIP}$ in column 7.\footnote{We fine-tune CLIP on multimodal downstream tasks (VQAv2, SNLI-VE, and HM) by applying a classifier on the concatenation of the two output vectors from its image and text encoders (details in supplemental).} Comparing column 3 vs 7, we see that $\mathrm{FLAVA}_\mathrm{C}$ outperforms it in all vision, language, and multimodal domains. This can be attributed to mostly two factors: different model details of FLAVA (\eg 768 text encoder hidden size instead of 512) and performing global back-propagation across all GPU workers as mentioned in Sec.~\ref{sec:multimodal}. In a more detailed analysis, we find that the latter improves our macro average over vision, NLP, and multi-modal tasks by +1.65\% with only minor additional computation overhead, indicating the global back-propagation implementation in contrastive loss are critical to effective pretraining.

\myparagraph{MMM and ITM objectives benefit multimodal tasks.} Next, we include the other multimodal objectives from Sec.~\ref{sec:multimodal} into our pretraining, using $L_{\mathrm{MMM}}$ and $L_{\mathrm{ITM}}$ along with $L_{\mathrm{GC}}$. The results are denoted as $\mathrm{FLAVA}_\mathrm{MM}$ in Table \ref{tab:ablations} column 4. Compared to $\mathrm{FLAVA}_\mathrm{C}$ with only the contrastive loss $L_{\mathrm{GC}}$ (column 3 vs 4), this setting improves multimodal average score by +2.86\%, NLP average score by +9\%, and also vision average score slightly by +0.3\%.

We additionally compare $\mathrm{FLAVA}_\mathrm{MM}$ with two other baseline settings -- the FLAVA model trained with only unimodal MIM or MLM losses in Sec.~\ref{sec:unimodal}, respectively over the images or the text in PMD. These two baselines are shown in Table~\ref{tab:ablations} column 1 and 2, which are largely outperformed by $\mathrm{FLAVA}_\mathrm{MM}$. These results indicate that the combined multimodal objectives (contrastive, MMM, ITM) allow FLAVA to learn powerful representations for both unimodal and multimodal downstream tasks.

\myparagraph{Joint unimodal \& multimodal pretraining helps NLP.} For the full FLAVA pretraining, we introduce unimodal image data from ImageNet-1k (IN-1k) and text data from CCNews and BookCorpus (BC). In this setting, we apply $\mathrm{FLAVA}_\mathrm{MM}$ losses on PMD data batches, MIM loss on IN-1k unimodal image data and MLM loss on CCNews text data, following Sec.~\ref{sec:unimodal}, shown in Table~\ref{tab:ablations} column 5. Comparing it to $\mathrm{FLAVA}_\mathrm{MM}$ in column 4 with only multimodal pretraining, this joint unimodal and multimodal pretraining improves the NLP average score from 74.22 to 75.55, which suggests that the additional text data from CCNews and BookCorpus benefits language understanding through the MLM objective.

However, we also observe from column 4 vs 5 that the macro average over all tasks decreases slightly. We suspect that this is because adding different tasks to the mix makes the optimization problem much harder, especially when the whole model is randomly initialized. Also, the round-robin sampling of tasks does not follow any particular curriculum to order the learning sequence of these tasks. Naturally, having some vision and language understanding is important before learning multimodal tasks, which motivates us to explore first leveraging unimodal pretraining before the joint training, as described below.

\myparagraph{Better image and text encoders via unimodal pretraining.} As detailed in Section \ref{sec:unimodal}, in order to leverage unimodal learning before joint training, we initialize the model from pretrained self-supervised weights for both vision and language encoders. For vision encoder, we initialize from an off-the-shelf DINO model pretrained on ImageNet-1k \cite{imagenet}. For the language encoder, we pretrain a ViT model with MLM loss on CCNews and BookCorpus datasets and use its model weights. Comparing column 5 vs 6, we observe pretrained encoders boost the performance of $\mathrm{FLAVA}$ on all tasks. We empirically find that initializing the vision encoder from a DINO self-supervised model gives better performance compared to a BEiT self-supervised model (see supplemental for additional details).

\subsection{Comparison to state-of-the-art models}
\label{sec:compare_sota}

We compare our full FLAVA model (Table~\ref{tab:ablations} column 6) with several state-of-the-art models on multimodal tasks, language tasks, and ImageNet linear evaluation, in Table~\ref{tab:sota_comp}. FLAVA largely outperforms previous multimodal approaches pretrained on public data (row 4 to 11) on both language and multimodal tasks and approaches the well-established BERT model on several GLUE tasks.

\begin{figure}[t]
\centering
\includegraphics[width=\columnwidth, keepaspectratio=True]{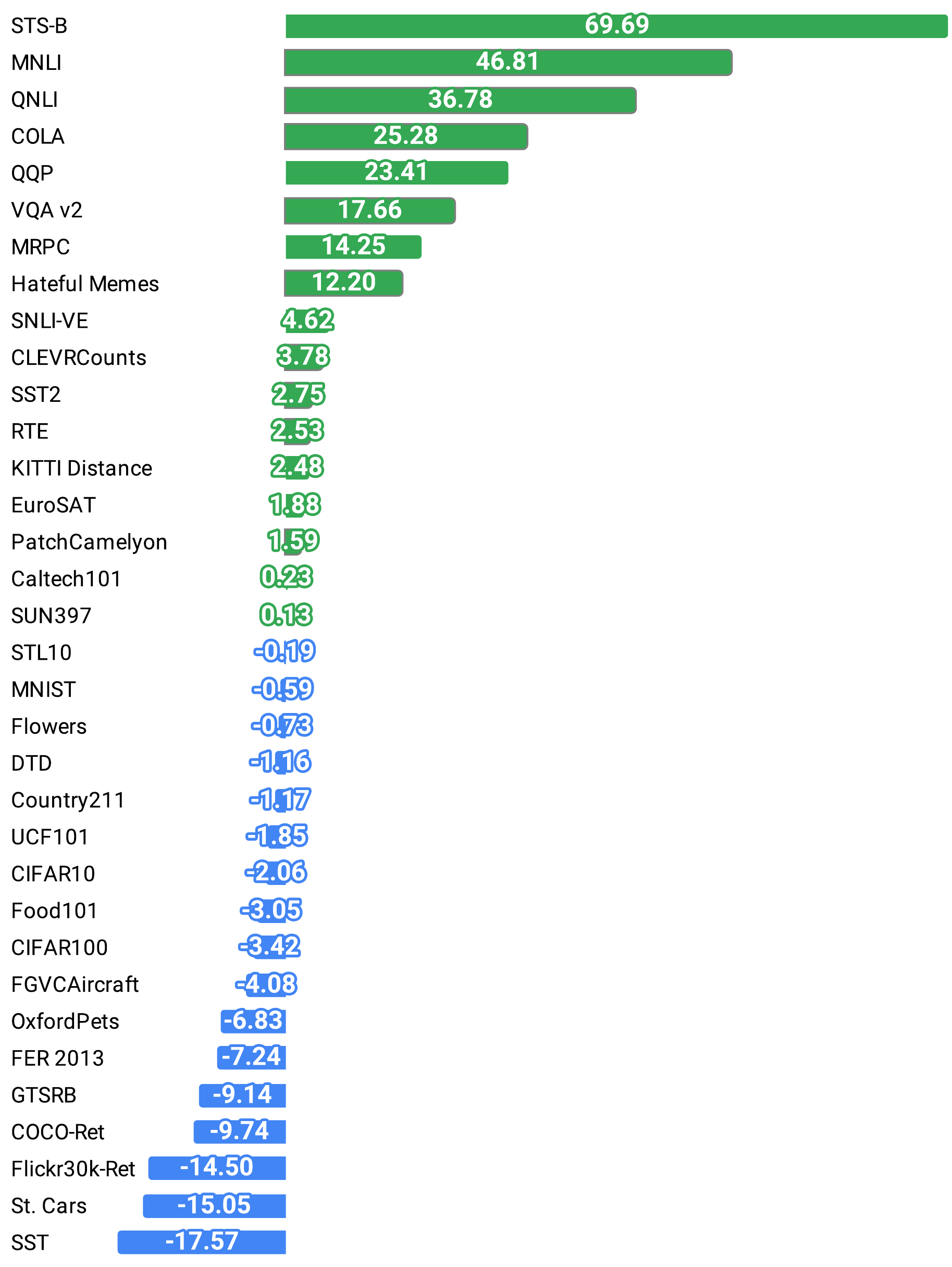}
\caption{The performance difference (relative, in \%) between FLAVA and the released CLIP-ViT-B/16 (400M) \cite{radford2021learning} on vision, language and multimodal tasks (positive means FLAVA is better).}
\vspace{-2.0em}
\label{fig:clip_vs_flava}
\end{figure}

FLAVA combines unimodal and multimodal losses and learns more generic representations which are transferable to vision, language, and multimodal tasks.  We evaluate the best released CLIP \cite{radford2021learning} ViT-B/16 model (pretrained on 400M image-text pairs in \cite{radford2021learning} with the same image encoder architecture as in FLAVA) on our task benchmark, shown in Table~\ref{tab:sota_comp} row 2. Compared to CLIP, we train FLAVA on just 70M data which is \textbf{$\sim$6x} smaller. In Fig.~\ref{fig:clip_vs_flava}, we observe that FLAVA works significantly better on language and multimodal tasks while slightly worse than CLIP on some vision-only tasks. In addition, we note that FLAVA outperforms the variant of the CLIP model pretrained only on the PMD dataset (Table~\ref{tab:sota_comp} row 10). Table~\ref{tab:ablations} further shows a breakdown analysis between our model (column 6) and the released CLIP ViT-B/16 (400M) model (column 8) and the CLIP trained on PMD (column 7).

FLAVA also has comparable performance to SimVLM \cite{wang2021simvlm} (Table~\ref{tab:sota_comp} row 3) on language tasks while underperforming it on multimodal tasks and ImageNet linear evaluation. FLAVA is pretrained using a much smaller dataset compared to 1.8B image-text pairs in \cite{wang2021simvlm}, and we anticipate that FLAVA's performance will further heavily improve as the pretraining dataset size increases.

\section{Conclusion}

In this work, we have presented a foundational vision and language alignment model that performs well on all three target modalities: 1) vision, 2) language, and 3) vision \& language. We introduced a novel set of objectives to achieve this goal and conducted experiments on a wide variety of 35 tasks to analyze the model's performance. FLAVA was trained on a corpus of publicly available datasets that is several orders of magnitude smaller than similar recent models, but still obtained better or competitive performance. Our work points the way forward towards generalized but open models that perform well on a wide variety of multimodal tasks.

\myparagraph{Broader impacts and limitations.} The models in this work are trained on public datasets widely used in the community. This enables reproducibility and we hope that our work will motivate others to compare models across a wide area of tasks and domains with the same data. However, like all natural data, these datasets have biases, potentially affecting our models. We partly mitigate this by combining several public datasets to increase the diversity and evaluating on an even larger set of target datasets. Still, further study is needed to identify and reduce potentially harmful biases. 

\small{\noindent\textbf{Acknowledgements.} We thank Devi Parikh for her support and advice on this project. We are grateful to Dmytro Okhonko, Hu Xu, Armen Aghajanyan, Po-Yao Huang, Min Xu, and Aleksandra Piktus for joint explorations of multimodal data. We thank Ning Zhang, Madian Khabsa, Sasha Sheng, and Naman Goyal for useful technical discussions; Karan Desai for providing access to RedCaps; Vaibhav Singh and others on the Google TPU team for TPU support; Shubho Sengupta, Armand Joulin, Brian O'Horo, Arthur Menezes for compute and storage support; and Ryan Jiang, Kushal Tirumala and Russ Howes for help running experiments.}

{\small
\bibliographystyle{ieee_fullname}
\bibliography{main}
}
\newpage

\twocolumn[{
\begin{center}
\Large 
\textbf{FLAVA: A Foundational Language And Vision Alignment Model}\\(Supplementary Material)
\par
\end{center}
\vspace{2em}
}]

\normalsize
\appendix

\counterwithin{figure}{section}
\counterwithin{table}{section}
\counterwithin{equation}{section}

\section{Hyperparameters and details of FLAVA}

\begin{table}[b]
\small
\begin{center}
\begin{tabular}{@{}lr@{}}
\toprule
\textbf{Hyperparameter} & \textbf{Value} \\
\midrule
\multicolumn{2}{c}{\textit{Image Encoder}} \\
\midrule
hidden size & 768 \\
number of heads & 12 \\
intermediate size & 3072 \\
number of layers & 12 \\
dropout prob. & 0 \\
patch size & $16\times16$ \\
input image size (pretraining) & $224\times224$ \\
input image size (VQAv2 fine-tuning) & $480\times480$ \\
input image size (all other evaluation) & $224\times224$ \\
\midrule
\multicolumn{2}{c}{\textit{Text Encoder}} \\
\midrule
hidden size & 768 \\
number of heads & 12 \\
intermediate size & 3072 \\
number of layers & 12 \\
dropout prob. & 0 \\
\midrule
\multicolumn{2}{c}{\textit{Multimodal Encoder}} \\
\midrule
hidden size & 768 \\
number of heads & 12 \\
intermediate size & 3072 \\
number of layers & 6 \\
dropout prob. & 0 \\
\midrule
\multicolumn{2}{c}{\textit{Others}} \\
\midrule
text vocabulary size & 30522 \\
image dVAE codebook size & 8192 \\
global contrastive loss projection dim & 512 \\
\midrule
\multicolumn{2}{c}{\textit{Training}} \\
\midrule
batch size & 8192 \\
learning rate & 1e-3 \\
learning schedule & \texttt{warmup\_cosine} \\
warmup updates & 10000\\
AdamW weight decay & 0.1 \\ 
AdamW $\beta_1$ & 0.9 \\
AdamW $\beta_2$ & 0.999 \\
\bottomrule
\end{tabular}
\end{center}
\vspace{-1.5em}
\caption{A summary of various hyperparameters in FLAVA.}
\label{tab:supp_hyper-parameters}
\end{table}

We summarize the hyperparameters in our FLAVA model in Table~\ref{tab:supp_hyper-parameters}. We also list the sampling probabilities of the datasets for joint pretraining in Table~\ref{tab:supp_sampling_ratio}, including PMD (multimodal paired image and text), ImageNet-1k (unimodal unpaired images), and CCNews \& BookCorpus (unimodal unpaired text).

\begin{table}[t]
\small
\begin{center}
\begin{tabular}{@{}l@{}r@{}}
\toprule
\textbf{Dataset} & \textbf{Sampling probability}
\\
\midrule
PMD & 0.70 \\
ImageNet-1k & 0.15 \\
CCNews \& BookCorpus & 0.15 \\
\bottomrule
\end{tabular}
\end{center}
\vspace{-1.5em}
\caption{Sampling probabilities of PMD (multimodal paired image and text), ImageNet-1k (unimodal unpaired images), and CCNews \& BookCorpus (unimodal unpaired text) for joint FLAVA pretraining on the three modalities.}
\label{tab:supp_sampling_ratio}
\vspace{-1em}
\end{table}

We find that a large batch size, a large weight decay, and a long warmup are helpful to stabilize training and prevent divergence under a large learning rate. Based on this finding, we performed a hyperparameter search based by monitoring the learning curve as well as monitoring the zero-shot image classification accuracy based on the image-text contrastive loss on using the text templates from CLIP \cite{radford2021learning} to obtain the hyperparameters above.

\section{Training and evaluation details}
\subsection{Pretraining details}

\myparagraph{Language encoder pretraining.} We follow RoBERTa$_{base}$ pretraining hyperparameters to train our pre-norm ViT-based text encoder \cite{liu2019roberta}. Specifically, we pretrain our text encoder using masked language modeling (MLM) \cite{devlin2018bert} on CCNews and BookCorpus for 125K iterations with a batch size of 2048 and a learning rate of 5e-4. We pick the best checkpoint based on the MLM loss without any further hyperparameter sweeps over RoBERTa's default configuration.

\myparagraph{Vision encoder pretraining.} We pretrain the image encoder in FLAVA on the ImageNet-1k dataset following either BEiT \cite{bao2021beit} or DINO \cite{caron2021emerging}. When pretraining a ViT-B/16 image encoder with BEiT, we adopt the hyperparameters and training details in \cite{bao2021beit} with a masked image modeling loss by predicting the dVAE visual tokens of the masked image patches. We also follow the training protocols in \cite{caron2021emerging} to pretrain a DINO ViT-B/16 model as our image encoder. As discussed in Sec.~\ref{sec:supp_ablation}, we empirically find that the DINO-pretrained image encoder gives better final performance.

\myparagraph{Full FLAVA pretraining.} We pretrain jointly on the unimodal and multimodal datasets, following the sampling probabilities of these datasets as provided in Table~\ref{tab:supp_sampling_ratio}. Specifically, for each update, we pick a dataset based on its sampling probability and obtain a complete batch from it. In all our ablations, we use a training schedule such that the PMD dataset is sampled for a total of 150K iterations. We monitor the zero-shot accuracy on ImageNet classification \cite{imagenet} every 8K updates and select the best checkpoint based on the ImageNet-1k zero-shot accuracy. We follow \cite{radford2021learning} to calculate the zero-shot accuracy.

\subsection{Vision, language and multimodal evaluation}

We evaluate the pretrained FLAVA model across a broad range of vision, natural language, and multimodal tasks. We discuss our evaluation details of these tasks below.

\myparagraph{Linear probing on vision tasks.} We perform linear probe evaluations on the datasets by closely following the setup described in \cite{radford2021learning}. We extract image features from the final layer of the image encoder (before the multi-modal encoder) and train a logistic regression classifier (\texttt{L-BFGS} implementation from \cite{scikit-learn}) on the extracted image features. We follow the hyperparameters similar to \cite{radford2021learning} : 1000 iterations, logistic regression $\lambda$ parameter sweep from 1e-6 to 1e6.

\myparagraph{Fine-tuning on NLP tasks.} For NLP tasks, we finetune the language encoder end to end for all the GLUE tasks. We add a classification head on top of the language encoder for all the tasks, except for the STS-B task, where we use a regression head. The hyperparameters we use for finetuning follow the setup of RoBERTa\cite{liu2019roberta}.\footnote{We follow hyperparameters used in \href{https://github.com/pytorch/fairseq/tree/main/examples/roberta}{FairSeq RoBERTa repo} for finetuning on GLUE tasks without any further sweeping.}

\myparagraph{Fine-tuning on multimodal VQA, SNLI-VE, and Hateful Memes.} We adopt the following settings when fine-tuning on VQA, SNLI-VE, and Hateful Memes, adding a 2-layer classifier head with a hidden dimension of 1536 on top of $\mathbf{h}_{\mathrm{CLS},M}$ from the multimodal encoder (corresponding to \texttt{[CLS\_M]}). For VQAv2, we use 1e-4 learning rate, 44000 updates, and an input image size of $480\times 480$. For SNLI-VE and Hateful Memes, we use 1e-5 learning rate, a total iteration number of 24000, and an input image size of $224\times 224$ (we use the OCR tokens extracted from the images as the textual input for Hateful Memes). On all these three tasks, we use the AdamW optimizer with a batch size of 256, 1e-2 weight decay, and 2000 warm-up iterations followed by a cosine decay schedule.

We use the same approach above to also evaluate the CLIP model on VQAv2, SNLI-VE, and Hateful Memes datasets. Since CLIP does not have a multimodal encoder, we concatenate the image feature vector from its image encoder and the text feature vector from its text encoder, apply a 2-layer classifier head (with the same hidden dimension of 1536) over the concatenated feature, and finetune the model following the same hyperparameters as for FLAVA.

\myparagraph{Zero-shot multimodal text and image retrieval.} We also evaluate the FLAVA model on the multimodal zero-shot retrieval tasks over the Flickr30K and COCO datasets, where the model needs to select a text caption based on a query image or select an image based on a query caption. We use the cosine similarities between the image and text feature computed in the global contrastive loss in FLAVA as the matching scores between the image and text modalities. Then, the text caption (or image) with the highest matching score to the query is retrieved. Similarly, we also evaluate the zero-shot text and image retrieval performance of the CLIP model using the cosine similarities between its image and text features.

\section{Additional ablations and analyses}
\label{sec:supp_ablation}

\begin{table*}[ht]
\centering
\scriptsize
\begin{tabular}{lccccccccccccc|c}
\toprule
&  \multicolumn{1}{c}{MIM} & \multicolumn{1}{c}{MLM} &
\multicolumn{4}{c}{$\mathrm{FLAVA}_\mathrm{C}$} & \multicolumn{3}{c}{$\mathrm{FLAVA}_\mathrm{MM}$}  & \multicolumn{3}{c}{$\mathrm{FLAVA}$} & \multicolumn{1}{c|}{CLIP} & \multicolumn{1}{c}{CLIP} \\
\cmidrule(r){2-2}
\cmidrule(r){3-3}
\cmidrule(r){4-7}
\cmidrule(r){8-10}
\cmidrule(r){11-13}
\cmidrule(r){14-14}
\cmidrule(r){15-15}
&   &  & \shortstack{local\\contrastive} &  &  \shortstack{BEiT\\init.} & \shortstack{DINO\\init.} &  & \shortstack{BEiT\\init.} & \shortstack{DINO\\init.} & & \shortstack{BEiT\\init.} & \shortstack{DINO\\init.} & &  \\
& \small\texttt{1} & \small\texttt{2} & \small\texttt{3} & \small\texttt{4} & \small\texttt{5} & \small\texttt{6} & \small\texttt{7} & \small\texttt{8} & \small\texttt{9} & \small\texttt{10} & \small\texttt{11} & \small\texttt{12} & \small\texttt{13} & \small\texttt{14} \\
\cmidrule(r){2-2}
\cmidrule(r){3-3}
\cmidrule(r){4-7}
\cmidrule(r){8-10}
\cmidrule(r){11-13}
\cmidrule(r){14-14}
\cmidrule(r){15-15}
Datasets &  PMD & PMD & \multicolumn{4}{c}{PMD} & \multicolumn{3}{c}{PMD} & \multicolumn{3}{c}{PMD+IN-1k+CCNews+BC} & PMD & 400M\cite{radford2021learning} \\
\midrule 
MNLI & -- & 73.22 & 70.65 & 70.99 & 74.12 & 74.23 & 76.82 & 78.59 & 78.74 & 78.06 & \underline{\textbf{80.96}} & 80.32 & 32.84 & 33.52 \\
COLA & -- & 39.55 & 9.76 & 17.58 & 15.30 & 14.92 & 38.97 & 39.41 & 45.04 & 44.22 & 44.52 & \underline{\textbf{50.65}} & 11.02 & 25.37 \\
MRPC & -- & 73.24 & 73.20 & 76.31 & 74.28 & 73.50 & 79.14 & 79.30 & 80.66 & 78.90 & \underline{\textbf{85.96}} & 84.16 & 68.74 & 69.91 \\
QQP & -- & 86.68 & 85.08 & 85.94 & 87.29 & 87.02 & 88.48 & 88.52 & 88.82 & 88.60 & \underline{\textbf{89.27}} & 88.74 & 59.16 & 65.33 \\
SST-2 & -- & 87.96 & 85.78 & 86.47 & 88.30 & 89.22 & 89.33 & 91.51 & 90.02 & 90.14 & \underline{\textbf{91.74}} & 90.94 & 83.49 & 88.19 \\
QNLI & -- & 82.32 & 70.25 & 71.85 & 80.67 & 80.93 & 84.77 & 86.05 & 86.23 & 86.40 & \underline{\textbf{88.52}} & 87.31 & 49.46 & 50.54 \\
RTE & -- & 50.54 & 49.10 & 51.99 & 52.71 & 49.82 & 51.99 & \underline{\textbf{57.76}} & 50.90 & 54.87 & 51.62 & \underline{\textbf{57.76}} & 53.07 & 55.23 \\
STS-B & -- & 78.89 & 60.08 & 57.28 & 76.93 & 76.17 & 84.29 & \underline{\textbf{86.70}} & 85.86 & 83.21 & 86.64 & 85.67 & 13.70 & 15.98 \\
\midrule
\textbf{NLP Avg.} & -- & 71.55 & 62.99 & 64.80 & 68.70 & 68.22 & 74.22 & 75.98 & 75.78 & 75.55 & 77.40 & \underline{\textbf{78.19}} & 46.44 & 50.51 \\
\midrule
ImageNet & 41.79 & -- & 70.64 & 74.09 & 74.07 & 75.87 & 74.34 & 74.37 & \textbf{76.23} & 73.49 & 74.59 & 75.54 & 72.95 & \underline{80.20} \\
Food101 & 53.30 & -- & 85.02 & 87.77 & 88.04 & \textbf{88.94} & 87.53 & 87.82 & 88.88 & 87.39 & 88.02 & 88.51 & 85.49 & \underline{91.56} \\
CIFAR10 & 76.20 & -- & 91.74 & \textbf{93.44} & 91.65 & 92.49 & 92.37 & 91.17 & 92.29 & 92.63 & 91.91 & 92.87 & 91.25 & \underline{94.93} \\
CIFAR100 & 55.57 & -- & 73.54 & \textbf{78.37} & 74.58 & 76.32 & 78.01 & 74.76 & 76.97 & 76.49 & 75.29 & 77.68 & 74.40 & \underline{81.10} \\
Cars & 14.71 & -- & 60.86 & \textbf{72.12} & 69.92 & 71.83 & 72.07 & 69.44 & 71.84 & 66.81 & 69.44 & 70.87 & 62.84 & \underline{85.92} \\
Aircraft & 13.83 & -- & 42.96 & \textbf{49.74} & 46.11 & 49.17 & 48.90 & 44.73 & 48.63 & 44.73 & 45.81 & 47.31 & 40.02 & \underline{51.40} \\
DTD & 55.53 & -- & 73.51 & 76.86 & 76.97 & \textbf{77.77} & 76.91 & 75.80 & 77.18 & 75.80 & 76.54 & 77.29 & 73.40 & \underline{78.46} \\
Pets & 34.48 & -- & 80.10 & 84.98 & 84.63 & 86.26 & 84.93 & 84.55 & \textbf{86.75} & 82.77 & 84.60 & 84.82 & 79.61 & \underline{91.66} \\
Caltech101 & 67.36 & -- & 92.98 & 94.91 & 94.95 & \textbf{\underline{95.94}} & 95.32 & 95.46 & 95.45 & 94.95 & 94.89 & 95.74 & 93.76 & 95.51 \\
Flowers & 67.23 & -- & 94.42 & 96.36 & 96.08 & \textbf{96.86} & 96.39 & 96.03 & 96.49 & 95.58 & 96.34 & 96.37 & 94.94 & \underline{97.12} \\
MNIST & 96.40 & -- & 97.75 & 98.39 & 98.28 & 98.49 & 98.58 & 97.94 & 98.38 & \textbf{98.70} & 98.38 & 98.42 & 97.38 & \underline{99.01} \\
STL10 & 80.12 & -- & 97.52 & 98.06 & 98.71 & 98.75 & 98.31 & 98.50 & \textbf{98.94} & 98.32 & 98.55 & 98.89 & 97.29 & \underline{99.09} \\
EuroSAT & 95.48 & -- & 95.76 & 97.00 & 97.04 & 97.24 & 96.98 & 97.36 & 96.72 & 97.04 & \textbf{\underline{97.40}} & 97.26 & 95.70 & 95.38 \\
GTSRB & 63.14 & -- & 73.81 & 78.92 & 74.76 & 79.27 & 77.93 & 76.13 & 79.01 & 77.71 & 76.96 & \textbf{79.46} & 76.34 & \underline{88.61} \\
KITTI & 86.03 & -- & 87.77 & 87.83 & 89.04 & 88.03 & 88.84 & \textbf{\underline{89.77}} & 89.71 & 88.70 & 88.57 & 89.04 & 84.89 & 86.56 \\
PCAM & 85.10 & -- & \textbf{\underline{86.04}} & 85.02 & 85.09 & 85.25 & 85.51 & 85.29 & 85.27 & 85.72 & 84.84 & 85.31 & 83.99 & 83.72 \\
UCF101 & 46.34 & -- & 77.82 & 82.69 & 80.60 & 82.90 & 82.90 & 81.52 & \textbf{83.40} & 81.42 & 81.60 & 83.32 & 77.85 & \underline{85.17} \\
CLEVR & 61.51 & -- & 73.86 & 79.35 & 80.24 & 79.84 & \textbf{\underline{81.66}} & 80.96 & 79.81 & 80.62 & 80.88 & 79.66 & 73.64 & 75.89 \\
FER 2013 & 50.98 & -- & 57.40 & 59.96 & 60.91 & 60.30 & 60.87 & 60.34 & \textbf{61.12} & 58.99 & 60.43 & \textbf{61.12} & 57.04 & \underline{68.36} \\
SUN397 & 52.45 & -- & 79.43 & 81.27 & 81.96 & \textbf{\underline{82.75}} & 81.41 & 81.99 & 82.16 & 81.05 & 81.76 & 82.17 & 79.96 & 82.05 \\
SST & 57.77 & -- & 58.65 & 56.67 & 58.05 & 58.98 & \textbf{59.25} & 56.29 & 57.17 & 56.40 & 56.12 & 57.11 & 56.84 & \underline{74.68} \\
Country211 & 8.87 & -- & 22.98 & 27.27 & 26.87 & 27.84 & 26.75 & 26.64 & 27.69 & 27.01 & 27.28 & \textbf{28.92} & 25.12 & \underline{30.10} \\
\midrule
\textbf{Vision Avg.} & 57.46 & -- & 76.12 & 79.14 & 78.57 & \textbf{79.59} & 79.35 & 78.49 & 79.55 & 78.29 & 78.65 & 79.44 & 76.12 & \underline{82.57} \\
\midrule
VQAv2 & -- & -- & 65.82 & 67.13 & 66.98 & 68.34 & 71.69 & 73.14 & \textbf{\underline{73.75}} & 71.29 & 72.23 & 72.49 & 59.81 & 54.83 \\
SNLI-VE & -- & -- & 74.03 & 73.27 & 74.37 & 73.59 & 78.36 & \textbf{\underline{79.05}} & 79.01 & 78.14 & 78.49 & 78.89 & 73.53 & 74.27 \\
Hateful Memes & -- & -- & 59.31 & 55.58 & 63.20 & 59.65 & 70.72 & 69.61 & \textbf{\underline{79.69}} & 77.45 & 74.10 & 76.09 & 56.59 & 63.93 \\
Flickr30K TR R@1 & -- & -- & 68.80 & 68.30 & 64.90 & 70.80 & 69.30 & \textbf{71.00} & 69.80 & 64.50 & 69.50 & 67.70 & 60.90 & \underline{82.20} \\
Flickr30K TR R@5 & -- & -- & 91.80 & 93.50 & 92.20 & 92.90 & 92.90 & 91.80 & 92.00 & 90.30 & 93.00 & \textbf{94.00} & 88.90 & \underline{96.60} \\
Flickr30K IR R@1 & -- & -- & 59.24 & 60.56 & 63.14 & 65.06 & 63.16 & 64.60 & 64.84 & 60.04 & 63.78 & \textbf{\underline{65.22}} & 56.48 & 62.08 \\
Flickr30K IR R@5 & -- & -- & 84.58 & 86.68 & 87.94 & 89.32 & 87.70 & 87.98 & 88.94 & 86.46 & 87.94 & \textbf{\underline{89.38}} & 83.60 & 85.68 \\
COCO TR R@1 & -- & -- & \textbf{48.28} & 43.08 & 44.00 & 45.06 & 43.48 & 42.44 & 44.62 & 39.88 & 42.24 & 42.74 & 37.12 & \underline{52.48} \\
COCO TR R@5 & -- & -- & 76.96 & 75.82 & 75.90 & 77.04 & 76.76 & 75.66 & \textbf{\underline{77.34}} & 72.84 & 75.38 & 76.76 & 69.48 & 76.68 \\
COCO IR R@1 & -- & -- & 37.34 & 37.59 & 38.28 & \textbf{\underline{39.20}} & 38.46 & 37.54 & 38.99 & 34.95 & 37.89 & 38.38 & 33.29 & 33.07 \\
COCO IR R@5 & -- & -- & 64.41 & 67.28 & 67.29 & \textbf{\underline{68.20}} & 67.68 & 66.71 & 67.70 & 64.63 & 66.96 & 67.47 & 62.47 & 58.37 \\
\midrule
\textbf{Multimodal Avg.} & -- & -- & 66.42 & 66.25 & 67.11 & 68.11 & 69.11 & 69.05 & \textbf{\underline{70.61}} & 67.32 & 69.23 & 69.92 & 62.02 & 67.29 \\
\midrule
\textbf{Macro Avg.} & 28.73 & 35.77 & 69.55 & 71.97 & 73.64 & 73.91 & 76.79 & 77.24 & 77.67 & 76.92 & 78.03 & \textbf{\underline{78.82}} & 61.28 & 66.54 \\
\bottomrule
\end{tabular}
\caption{\textbf{Comparing our full FLAVA pretraining with other settings} (similar to Table 4 in the main paper) with additional ablations (see Sec.~\ref{sec:supp_ablation} for details). The overall best result is \underline{underlined} while \textbf{bold} signifies the best on public data (PMD and unimodal).}
\label{tab:ablations_appendix}
\end{table*}

\myparagraph{Unimodal-pretrained vision encoders.} We experiment with initializing our model from different pretrained vision encoders (while keeping the language encoder the same). We study two different self-supervised ViT-B/16 models trained on ImageNet-1k: i) BEiT and ii) DINO. Under three FLAVA pretraining settings, $\mathrm{FLAVA}_\mathrm{C}$, $\mathrm{FLAVA}_\mathrm{MM}$ and $\mathrm{FLAVA}$ (full pretraining), initializing from any of the two pretrained vision encoders (along with pretrained language encoders) leads to significant improvement in all tasks. In Table~\ref{tab:ablations_appendix}, comparing columns 5 vs 6, 8 vs 9, and 11 vs 12 between BEiT and DINO initialization, DINO gives better performance on vision and multimodal tasks. On NLP tasks, the results are mixed and comparable, as the language encoder is initialized from the same pretrained weights. 
\myparagraph{Global vs. local contrastive losses.} In our FLAVA model, we apply a global contrastive loss, where the image and text features are gathered across GPUs and the loss is back-propagated through the gathering operation to all GPUs. This is in contrast with the implementation in \cite{ilharco_gabriel_2021_5143773}, where the loss is only back-propagated to local features from the same GPU. It can be seen from Table~\ref{tab:ablations_appendix} (columns 3 vs 4) that the global contrastive loss (column 4) leads to a noticeable gain in the average vision and NLP performance compared to its local contrastive counterpart and also provides a slight boost in multimodal performance. 

\myparagraph{Observations on SST and VQA.} Some of our vision tasks involve classifying an image using the text written on the image pixels, and require the model to perform OCR to read text from images. For example, in the SST task in Table~\ref{tab:ablations_appendix} (which is also evaluated as an image classification task in \cite{radford2021learning}), the model is asked to classify the sentiment of a natural language sentence by printing the sentence words onto an image and providing the image pixels to the model. It can be seen from Table~\ref{tab:ablations_appendix} that our FLAVA model does not perform well on this SST task, which we believe is mostly because our PMD dataset does not contain enough scene text information for the model to acquire text reading ability from images. We note that the CLIP model pretrained on PMD (column 13) has a similar lower performance on SST than the variant pretrained on 400M image-text pairs (column 14), and we anticipate that FLAVA will also be able to perform scene text reading when pretrained on a larger dataset with enough scene text information.

Our FLAVA model reaches a final accuracy of 72.49 on the VQAv2 dataset. While this accuracy is below the state-of-the-art on VQAv2, we note that this is a reasonable performance given the amount of data used in FLAVA pretraining. Recent models such as SimVLM \cite{wang2021simvlm} often use a much larger dataset (\eg 1.8B image-text pairs \cite{wang2021simvlm}), and we believe more pretraining data will also benefit FLAVA.

\section{Architectural differences between FLAVA and CLIP encoders}

\begin{table}[ht!]
\resizebox{\columnwidth}{!}{
\begin{tabular}{lccc|c}
\toprule
method & \shortstack{Vision \\Avg.} & \shortstack{NLP \\Avg.} & \shortstack{Multi-modal \\Avg.} & \shortstack{Macro \\Avg.} \\
\midrule
\texttt{1} CLIP (PMD) & 76.12 & 46.44 & 62.02 & 61.52 \\
\texttt{2} arch optimizations & 76.12 & 62.99 & 66.42 & 68.51   \\
\midrule
\hspace{3\tabcolsep} $\Delta$ & +0.00 & +16.55 & +4.40 & +6.99 \\\bottomrule
\end{tabular}}
\vspace{-0.5em}
\caption{Comparing our FLAVA image and text encoders to the original CLIP when trained under same settings on PMD.}
\label{tab:robust_clip}
\vspace{-0.5em}
\end{table}

FLAVA and CLIP \cite{radford2021learning} use transformers \cite{vaswani2017attention} as the image and text encoders in their comparable variations (column 3, FLAVA$_{C}$-local contrastive and column 13, CLIP-ViT-B/16 in Table \ref{tab:ablations_appendix}). Compared to CLIP which uses a text vocabulary of size 49152, in FLAVA we use BERT's text vocabulary with a size of 30522. CLIP uses lower-cased byte pair encoding similar to \cite{sennrich2015neural,radford2018improving} whereas we use BERT's tokenizer from \cite{wolf2020huggingfaces} to tokenize the text. Furthermore, we use a hidden size of 768 instead of 512 and use the ViT architecture (based on the implementation in Hugging Face \cite{wolf2020huggingfaces}) instead of the GPT-style transformer architecture in CLIP for both text and image encoders \cite{xiong2020layer}. Table~\ref{tab:robust_clip} shows the comparison of macro averages for the three domains between the original CLIP architecture and our optimized FLAVA architecture trained on PMD under the same settings with local contrastive loss (corresponding to columns 13 and 3 in Table \ref{tab:ablations_appendix}, respectively). A comparison between rows 1 and 2 in Table~\ref{tab:robust_clip} shows that our architecture optimizations help achieve a better macro average overall.

\end{document}